\documentclass[conference]{IEEEtran}

\usepackage{cite}
\usepackage{amsmath,amssymb,amsfonts}
\usepackage{graphicx}
\usepackage{textcomp}
\usepackage[table]{xcolor}
\usepackage{booktabs}
\usepackage{multirow}
\usepackage{array}
\usepackage{rotating}
\definecolor{NAWIFBlue}{RGB}{225,235,252}

\def\BibTeX{{\rm B\kern-.05em{\sc i\kern-.025em b}\kern-.08em
    T\kern-.1667em\lower.7ex\hbox{E}\kern-.125emX}}

\title{Negation-Aware Weighted Information Fusion for Reliable Evidential Reasoning}

\author{
\IEEEauthorblockN{
Qiying Hu\IEEEauthorrefmark{2},
Rui Sun\IEEEauthorrefmark{3}
Yingying Liang\IEEEauthorrefmark{4},
Qianli Zhou\IEEEauthorrefmark{1}, and
Witold Pedrycz\IEEEauthorrefmark{5},~\textit{Life Fellow, IEEE}}
\IEEEauthorblockA{\IEEEauthorrefmark{1}Northwestern Polytechnical University}
\IEEEauthorblockA{\IEEEauthorrefmark{2}University of Electronic Science and Technology of China} 
\IEEEauthorblockA{\IEEEauthorrefmark{3}Tsinghua University} 
\IEEEauthorblockA{\IEEEauthorrefmark{4}Hebei University of Technology} 
\IEEEauthorblockA{\IEEEauthorrefmark{5}University of Alberta}
}

\begin{document}
\maketitle

\begin{abstract}
Dempster--Shafer theory (DST) provides an attractive framework for
multi-source information fusion. However, its
effectiveness critically depends on how a BPA is constructed for each
source. Existing methods often rely on heuristic mappings or restrict belief assignment to singleton hypotheses and the full set, thereby underutilizing the expressive belief structure of BPAs for representing ambiguity.
This paper proposes Negation-Aware Weighted Information Fusion (NAWIF).
First, NAWIF transforms the class-wise possibility distribution of each
source into a set of negation-aware simple BPAs, which encode class
exclusion through complementary propositions and retain residual
uncertainty in the full set. Their combination produces a
closed-form BPA over composite hypotheses, preserving
ambiguity among the classes not excluded by the source. It further performs source-specific BPA discounting based on its reliability and uncertainty, suppressing the impact of unreliable or uninformative sources.
Experiments on benchmark pattern recognition datasets demonstrate the effectiveness of NAWIF, which achieves state-of-the-art average performance with particularly pronounced advantages under noise disturbance. 
\end{abstract}


\section{Introduction}
\label{sec:introduction}

Uncertainty reasoning~\cite{huang2025towards,he2026survey} aims to derive reliable decisions by aggregating heterogeneous information sources that may be incomplete, imprecise, or mutually conflicting. Dempster--Shafer theory (DST)~\cite{shafer1976mathematical} provides a natural and promising
framework for this purpose. It represents each source by a basic
probability assignment (BPA) and combines the resulting evidence across
sources. Unlike precise probabilistic modeling, a BPA can allocate
belief to both singleton and composite hypotheses, explicitly retaining
partial ignorance and ambiguity until fusion~\cite{hu2024evidence,li2026novel}. Hence, it has been widely applied to pattern classification~\cite{xiao2025generalized,xiao2026adaptive}, multi-source information fusion~\cite{chen2026trustworthyqenn,xiao2024complex}, reliability assessment~\cite{chen2024risk,he2024generalized}, and decision-making~\cite{gao2025information,su2026novel}.

DST-based reasoning has been widely applied to pattern recognition,
where decisions often rely on uncertain or incomplete observations.
Denoeux's evidential $k$-nearest neighbor (EKNN)
classifier~\cite{denoeux1995k,gong2023distributed} represents each neighbor by a
distance-dependent simple BPA and combines these BPAs using
Dempster's rule. Deng \textit{et al.}~\cite{deng2022random,deng2024random}
develops random permutation set reasoning (RPSR), which constructs
permutation mass functions  and
incorporates hypothesis ordering and source reliability into reasoning.
Xu \textit{et al.}~\cite{xu2013new,luo2026attribute} proposed an attribute-fusion-based
classifier that fits class-conditional Gaussian models for each
attribute, transforms their output into BPAs with nested structure, and combines evidence across attributes. These approaches highlight that reliable evidential reasoning fundamentally relies on representing each information source with a meaningful and trustworthy BPA.

More recently, evidence theory has been further integrated into deep neural networks, extending evidential modeling from post-hoc decision fusion to end-to-end representation learning. Sensoy \textit{et al.}~\cite{sensoy2018evidential,gao2025comprehensive} introduced deep evidential learning, where neural networks learn evidence for Dirichlet distributions over class probabilities to jointly support prediction and uncertainty estimation. Subsequent studies have incorporated evidential representations into diverse learning paradigms, including uncertainty-aware pseudo-labeling for semi-supervised learning and test time adaptation~\cite{mai2026softmax,fan2026evidential}, quantum evidential neural networks for reliable pattern classification~\cite{chen2026trustworthyqenn}, and evidence fusion for modeling cross-channel dependencies~\cite{zhan2025time}. Collectively, these developments demonstrate that evidential reasoning can serve not only as a decision-level fusion mechanism but also as an integral component of deep learning for explicitly representing uncertainty and integrating heterogeneous information.

Despite these advances, two limitations remain in existing BPA generation methods for pattern recognition. First, many approaches rely on heuristic allocation rules~\cite{xu2013new,deng2024random} or restrict belief assignment to singleton hypotheses and the full set~\cite{huang2023Evidential,gong2022sparse}. Under noisy or ambiguous observations, such restricted allocations may assign excessive mass to individual singletons, producing over-committed BPAs and exacerbating conflict in subsequent fusion. Second, the reliability and uncertainty of evidence can vary substantially across both sources and samples. Treating all sources as equally trustworthy may therefore allow unreliable or ambiguous evidence to exert disproportionate influence on the fusion result~\cite{xu2013new,deng2024random}. To address these limitations, we propose Negation-Aware Weighted Information Fusion (NAWIF), which comprises two complementary strategies. First, \emph{negation-aware BPA generation} incorporates negation information by assigning the negation of each class possibility to its complementary proposition and combining the resulting simple BPAs into a source-level BPA. Second, \emph{adaptive evidential discounting} jointly accounts for global source reliability and uncertainty to regulate each source's contribution, transferring discounted mass to the full set before fusion. Experiments on eight benchmark datasets demonstrate that NAWIF achieves the best average performance among both machine-learning and evidential reasoning-based classifiers, while attaining the highest Macro-F1 score on nearly all datasets under Gaussian perturbations. Further analysis shows that negation-aware BPA generation substantially reduces fusion conflict, leading to more robust and reliable evidence combination.

\section{Preliminaries}
Let $\Theta=\{\theta_1,\ldots,\theta_C\}$ denote the frame of
discernment (FoD), where the $C$ hypotheses are mutually exclusive and
exhaustive. Its power set is $ 2^\Theta=
\left\{
\emptyset,\,
\{\theta_i\},\,
\{\theta_i,\theta_j\},\,
\{\theta_i,\theta_j,\theta_k\},\,
\ldots,\,
\Theta
\right\},$
where the indices range over all valid distinct combinations. A basic probability
assignment (BPA) is a mapping $m:2^\Theta\rightarrow[0,1]$ satisfying
\begin{equation}
m(\emptyset)=0,
\qquad
\sum_{A\subseteq\Theta}m(A)=1,
\label{eq:bpa}
\end{equation}
where $\emptyset$ denotes the empty set. Each subset
$A\subseteq\Theta$ with $m(A)>0$ is referred to as a focal element.
By assigning mass to both singleton, composite hypotheses, and the full set, a BPA can
explicitly represent uncertainty, ambiguity, and partial ignorance. Given $K$ independent sources with BPAs
$m_1,\ldots,m_K$ defined on $\Theta$, their Dempster's rule of
combination (DRC) is given, for each nonempty $A\subseteq\Theta$, by
\begin{equation}
m_{\mathrm{fus}}(A)
=
\frac{1}{1-\kappa}
\sum_{\substack{B_1,\ldots,B_K\subseteq\Theta\\
\bigcap_{j=1}^{K}B_j=A}}
\prod_{j=1}^{K}m_j(B_j),
\label{eq:drc}
\end{equation}
where $B_j$ denotes a focal subset of the $j$-th source, and
$m_{\mathrm{fus}}$ is the combined BPA. The conflict coefficient is
defined as
\begin{equation}
\kappa
=
\sum_{\substack{B_1,\ldots,B_K\subseteq\Theta\\
\bigcap_{j=1}^{K}B_j=\emptyset}}
\prod_{j=1}^{K}m_j(B_j).
\label{eq:conflict}
\end{equation}
Here, $\kappa$ measures the total conflicting degree among the $K$
sources. For decision making, let $m_{\mathrm{fus}}$ denote the final fused BPA. The pignistic probability transformation (PPT) converts it into class probabilities:
\begin{equation}
\mathrm{BetP}(\theta_i)
=
\sum_{\substack{A\subseteq\Theta\\\theta_i\in A}}
\frac{m_{\mathrm{fus}}(A)}{|A|},
\qquad
\hat{\theta}
=
\arg\max_{\theta_i\in\Theta}\mathrm{BetP}(\theta_i),
\label{eq:betp}
\end{equation}
where $|A|$ denotes the cardinality of $A$, and $\hat{\theta}$ is the
final decision. Thus, DST-based reasoning represents source information as
BPAs, combines the BPAs in the belief space, and transforms the fused
evidence into a final decision.
\section{Proposed Method}
\label{sec:method}
\begin{figure}
    \centering
    \includegraphics[width=1.05\linewidth]{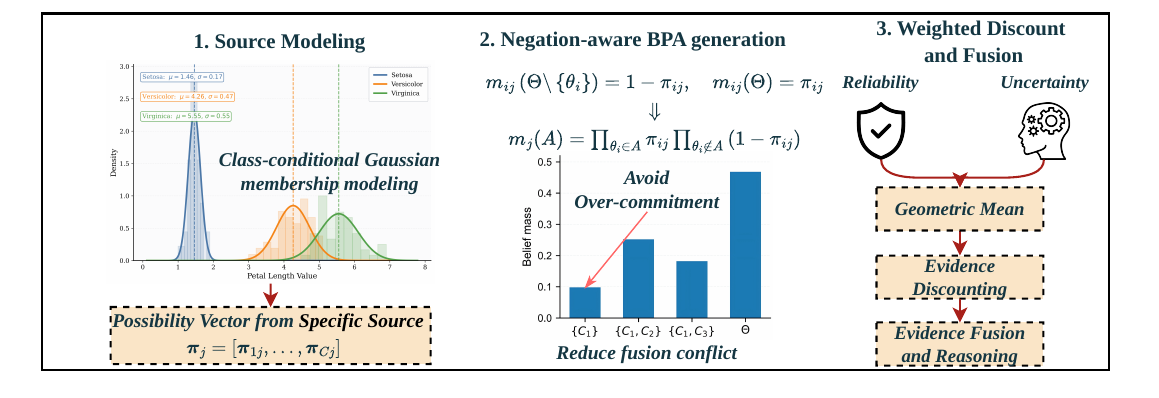}
    \caption{Overview of the proposed NAWIF.}
    \label{fig:placeholder}
\end{figure}
\subsection{Problem Formulation and Overall Framework}
\label{sec:framework}

Consider a multi-source pattern recognition problem with \(K\) information sources $\mathcal{S}=\{S_1,\ldots,S_K\}$ defined on a FoD $\Theta=\{\theta_1,\ldots,\theta_C\}$. For sample
$\mathbf{x}_n=[x_{n1},\ldots,x_{nK}]$, each feature $x_{nj}$ represents
a specific information source, which is transformed into a BPA
$m_{nj}$ and subsequently combined for final inference. As illustrated in Fig.~\ref{fig:placeholder}, NAWIF comprises three stages. First, each feature is treated as an information source and modeled using class-conditional Gaussian membership functions, yielding a class-wise possibility vector. Second, the possibility vector is transformed into a negation-aware BPA by assigning class incompatibility to the corresponding complementary proposition while retaining residual uncertainty on the full set. Combining the resulting simple BPAs yields a closed-form source-level BPA over composite focal elements, thereby explicitly representing ambiguity among plausible classes. Third, NAWIF determines an adaptive discounting factor for each source as the geometric mean of its global reliability and evidential information volume. The discounted BPAs are subsequently fused using DRC, followed by PPT for final decision making.

\subsection{Negation-Aware BPA Generation}

For each feature, we first establish a class-conditional membership function
model to characterize its support for different hypotheses.
Specifically, the membership function for hypothesis $\theta_i$ is
modeled by a Gaussian distribution:
\begin{equation}
f_{ij}(x)
=
\frac{1}{\sqrt{2\pi\sigma_{ij}^{2}}}
\exp\left(
-\frac{(x-\mu_{ij})^{2}}{2\sigma_{ij}^{2}}
\right),
\label{eq:membership}
\end{equation}
where $\mu_{ij}$ and $\sigma_{ij}$ denote the mean and standard
deviation estimated from the training samples belonging to
$\theta_i$ for feature $j$, respectively. Given a test observation
$x_{nj}$, the corresponding class-wise membership responses are
normalized to obtain the possibility vector
$\boldsymbol{\pi}_{nj}=[\pi_{1nj},\ldots,\pi_{Cnj}]$:
\begin{equation}
\pi_{inj}
=
\frac{f_{ij}(x_{nj})}
{\max_{c\in\{1,\ldots,C\}} f_{cj}(x_{nj})}.
\label{eq:normalization}
\end{equation}
$\pi_{inj}$ characterizes the compatibility between the observation
provided by feature $j$ and hypothesis $\theta_i$. We interpret $\pi_{inj}$ as both support for $\theta_i$ and
rejection of the remaining hypotheses. Accordingly, its complement
$1-\pi_{inj}$ represents support for these alternatives collectively,
which we associate with the complementary proposition
$\Theta\setminus\{\theta_i\}$. Based on this interpretation,
a simple BPA is constructed as
\begin{equation}
m_{inj}(A)
=
\begin{cases}
1-\pi_{inj}, & A=\Theta\setminus\{\theta_i\},\\
\pi_{inj},   & A=\Theta,\\
0,           & \mathrm{otherwise}.
\end{cases}
\label{eq:negative_bpa}
\end{equation}
Under DRC, obtaining a focal set $A$ requires retaining every
$\theta_i\in A$ while excluding every $\theta_i\notin A$. Hence,
the $C$ simple BPAs associated with feature $j$ are combined as
$m_{nj}=\bigoplus_{i=1}^{C}m_{inj}$, yielding, for every nonempty
$A\subseteq\Theta$,
\begin{equation}
m_{nj}(A)
=
\prod_{\theta_i\in A}\pi_{inj}
\prod_{\theta_i\notin A}(1-\pi_{inj}).
\label{eq:closedform}
\end{equation}

This compact expression admits an elegant interpretation: the belief
assigned to $A$ jointly reflects the plausibility of retaining all
classes in $A$ and the evidence for excluding all classes outside $A$,
thereby naturally representing ambiguity among the remaining plausible
hypotheses.

\subsection{Adaptive Evidential Discounting and Fusion}
\label{sec:weighted_fusion}

Since information sources may differ in both reliability and
uncertainty, we estimate the global reliability $G_j$ of each source
and the sample-specific evidential information volume $I_{nj}$ prior to
combination. Global reliability is estimated from the training set as
\begin{equation}
G_j
=
1-\frac{1}{2N_{\mathrm{tr}}}
\sum_{n=1}^{N_{\mathrm{tr}}}
\left\|
\mathbf{p}_{nj}-\mathbf{y}_n
\right\|_2^2,
\label{eq:global_rel}
\end{equation}
where $\mathbf{p}_{nj}$ is the probability distribution transformed from
the BPA generated by source $j$, and $\mathbf{y}_n$ denotes the
corresponding one-hot label. A larger \(G_j\) indicates better agreement between the predictions of source \(j\) and the ground-truth labels, reflecting higher source reliability.

The evidential information volume of \(m_{nj}\) is measured by
\begin{equation}
I_{nj}
=
1-m_{nj}(\Theta),
\label{eq:info_rel}
\end{equation}
which quantifies the proportion of belief committed to informative
subsets rather than complete ignorance. The two factors are combined
through their geometric mean,
$r_{nj}=\sqrt{G_j I_{nj}}$, and used to discount the corresponding BPA:
\begin{equation}
\widetilde m_{nj}(A)
=
\begin{cases}
r_{nj}m_{nj}(A), & A\neq\Theta,\\
r_{nj}m_{nj}(\Theta)+(1-r_{nj}), & A=\Theta.
\end{cases}
\label{eq:discount}
\end{equation}
This discounting mechanism adaptively regulates the contribution of each
source. Evidence with a large $r_{nj}$ is largely retained, whereas a
small $r_{nj}$ transfers more mass to $\Theta$, reducing the influence
of unreliable or uninformative evidence without favoring any specific
hypothesis. Finally, the discounted BPAs are fused using DRC via
\eqref{eq:drc} and transformed into probability distributions through
\eqref{eq:betp} for final decision making.

\section{Application of NAWIF in pattern recognition}

\begin{table*}[t]
\centering

\caption{
Classification performance (\%) comparison on clean datasets under five-fold cross-validation.
Each entry is reported as ACC/F1, and the best results are shown in bold.
}
\label{tab:overall_comparison}

\fontsize{7}{8}\selectfont
\setlength{\tabcolsep}{2.6pt}
\renewcommand{\arraystretch}{0.95}

\resizebox{\textwidth}{!}{%
\begin{tabular}{
@{}
c
l
c c c c c c c c
c c
@{}
}

\toprule

&
\textbf{Method}
& \textbf{Haberman}
& \textbf{BCO}
& \textbf{MGM}
& \textbf{Wine}
& \textbf{Diabetes}
& \textbf{Iris}
& \textbf{BCD}
& \textbf{Heart}
& \textbf{Avg.}
& \textbf{Avg. Rank}
\\

\midrule

\multirow{6}{*}{
\rotatebox{90}{
\fontsize{7.5}{8.2}\selectfont
\textbf{ML-based}
}
}

& BayesNet
& 72.07/56.51
& 96.70/96.39
& \textbf{83.15/82.96}
& 95.59/95.70
& 74.44/71.72
& 93.91/93.82
& 95.32/94.98
& 76.54/75.99
& 85.97/83.51
& 5.00/5.13
\\

& SVM
& 72.99/47.65
& 95.63/95.11
& 80.82/80.66
& 96.43/96.40
& \textbf{76.55}/71.98
& 95.52/95.49
& \textbf{97.14/96.91}
& 81.19/80.75
& 87.03/83.12
& 3.63/5.00
\\

& DT
& 65.43/55.41
& 93.75/93.03
& 75.91/75.61
& 92.78/92.85
& 70.34/67.29
& 94.57/94.54
& 93.29/92.82
& 74.52/74.11
& 82.57/80.71
& 10.25/9.56
\\

& k-NN
& 69.51/51.71
& 96.88/96.56
& 78.67/78.52
& 95.78/95.84
& 73.46/69.64
& \textbf{95.79/95.77}
& 96.70/96.44
& 80.16/79.77
& 85.87/83.03
& 5.25/5.63
\\

& AdaBoost
& 73.01/59.22
& 95.57/95.08
& 82.90/82.70
& 95.40/95.51
& 75.48/71.95
& 93.99/93.95
& 96.91/96.67
& 81.13/80.71
& 86.80/84.47
& 4.88/5.00
\\

& MLP
& 71.66/44.47
& 95.36/94.91
& 76.84/76.72
& 75.07/71.63
& 64.87/43.77
& 51.39/41.14
& 89.65/88.60
& 73.31/71.69
& 74.77/66.62
& 11.38/12.38
\\

\midrule

\multirow{6}{*}{
\rotatebox{90}{
\fontsize{7.3}{8.0}\selectfont
\textbf{ER-based}
}
}

& DRCM
& 61.46/54.73
& 95.97/95.61
& 79.88/79.86
& \textbf{97.38/97.46}
& 73.74/69.74
& 94.67/94.64
& 93.41/92.91
& 84.43/84.18
& 85.12/83.64
& 5.38/4.63
\\

& ER Rule
& 65.79/57.21
& 96.21/95.80
& 79.68/79.65
& 93.63/93.80
& 73.81/69.28
& 95.45/95.43
& 93.71/93.16
& 84.26/83.91
& 85.32/83.53
& 5.50/5.50
\\

& MRCM
& 60.84/53.08
& 95.02/94.41
& 78.43/78.34
& 91.01/91.29
& 73.14/68.42
& 93.89/93.85
& 93.43/92.81
& 82.81/82.48
& 83.57/81.84
& 8.75/8.63
\\

& EDM
& 60.84/53.08
& 94.28/93.54
& 78.38/78.28
& 88.83/89.18
& 72.72/67.67
& 93.39/93.31
& 93.40/92.76
& 82.49/82.17
& 83.04/81.25
& 9.94/9.69
\\

& BJSM
& 60.84/53.08
& 91.12/89.61
& 78.38/78.28
& 81.00/80.69
& 69.23/59.29
& 81.47/81.59
& 92.84/92.07
& 82.29/81.97
& 79.65/77.07
& 11.19/10.81
\\

& RPSRCA
& 61.30/53.65
& 96.25/95.84
& 80.76/80.69
& 95.80/95.91
& 73.70/69.33
& 86.78/86.59
& 93.42/92.82
& 81.54/81.26
& 83.69/82.01
& 6.88/6.44
\\

\rowcolor{NAWIFBlue}
&
\textbf{NAWIF}
& \textbf{75.20/62.95}
& \textbf{97.17/96.89}
& 80.99/80.90
& 96.95/97.06
& 74.30/\textbf{72.55}
& 95.56/95.53
& 93.94/93.46
& \textbf{84.64/84.40}
& \textbf{87.34/85.47}
& \textbf{2.38/2.00}
\\

\bottomrule

\end{tabular}
}

\vspace{1.2mm}

\raggedright
\fontsize{7.1}{8.0}\selectfont
Avg. Rank is calculated across all compared methods;
a lower value indicates better overall performance.

\end{table*}

\begin{table*}[t]
\centering

\caption{
Classification performance (\%) under Gaussian noise applied to both training and test features. Each entry reports ACC/F1, with the best result for each metric highlighted in bold.
}
\label{tab:overall_comparison_noisy}

\fontsize{7}{8}\selectfont
\setlength{\tabcolsep}{2.6pt}
\renewcommand{\arraystretch}{0.95}

\resizebox{\textwidth}{!}{%
\begin{tabular}{
@{}
c
l
c c c c c c c c
c c
@{}
}

\toprule

&
\textbf{Method}
& \textbf{Haberman}
& \textbf{BCO}
& \textbf{MGM}
& \textbf{Wine}
& \textbf{Diabetes}
& \textbf{Iris}
& \textbf{BCD}
& \textbf{Heart}
& \textbf{Avg.}
& \textbf{Avg. Rank}
\\

\midrule

\multirow{6}{*}{
\rotatebox{90}{
\fontsize{7.5}{8.2}\selectfont
\textbf{ML-based}
}
}

& BayesNet
& 70.73/52.04
& 89.95/88.75
& 69.49/69.33
& 52.90/52.40
& 60.33/52.92
& 56.06/55.33
& 75.17/72.75
& 66.69/66.03
& 67.66/63.69
& 8.50/9.50
\\

& SVM
& \textbf{73.23}/44.24
& 91.13/89.85
& 69.73/68.68
& 66.31/65.87
& \textbf{64.48}/51.54
& 61.27/59.66
& 81.82/79.45
& 73.14/71.98
& 72.64/66.41
& 3.25/6.50
\\

& DT
& 63.48/52.98
& 85.31/83.70
& 62.75/62.49
& 55.00/54.49
& 56.66/52.46
& 54.83/54.38
& 69.38/67.26
& 66.23/65.61
& 64.20/61.67
& 11.62/11.00
\\

& k-NN
& 70.29/52.63
& 90.71/89.31
& 68.43/68.22
& 62.61/62.10
& 60.75/52.21
& 60.78/59.88
& 77.52/74.41
& 72.70/71.93
& 70.47/66.34
& 6.62/7.88
\\

& AdaBoost
& 72.20/53.87
& 90.50/89.34
& 70.68/70.43
& 64.83/64.51
& 64.13/53.46
& 59.62/59.35
& 79.42/77.65
& 73.31/72.78
& 71.84/67.67
& 4.50/5.12
\\

& MLP
& 71.56/44.33
& 89.28/88.26
& 70.73/70.58
& 49.75/45.42
& 63.48/42.27
& 43.68/36.17
& 75.99/73.64
& 64.32/62.45
& 66.10/57.89
& 8.38/10.75
\\

\midrule

\multirow{7}{*}{
\rotatebox{90}{
\fontsize{7.3}{8.0}\selectfont
\textbf{ER-based}
}
}

& DRCM
& 60.21/53.81
& 92.05/91.12
& 69.24/69.12
& 67.53/67.47
& 59.98/57.16
& 63.49/\textbf{63.30}
& 80.27/78.67
& 74.09/73.71
& 70.86/69.29
& 4.62/2.88
\\

& ER Rule
& 60.40/54.03
& 91.28/90.14
& 69.78/69.68
& 65.83/65.62
& 60.45/57.13
& 62.68/61.60
& 80.37/78.64
& 74.55/74.14
& 70.67/68.87
& 4.00/3.25
\\

& MRCM
& 59.78/52.64
& 89.75/88.28
& 66.59/66.40
& 62.86/62.31
& 59.97/57.14
& 62.00/60.61
& 79.14/77.34
& 72.26/71.81
& 69.04/67.07
& 7.94/6.75
\\

& EDM
& 59.78/52.64
& 88.41/86.61
& 66.23/66.03
& 60.92/60.13
& 60.02/57.10
& 60.37/58.34
& 78.88/77.01
& 71.45/70.95
& 68.26/66.10
& 9.06/8.44
\\

& BJSM
& 59.78/52.64
& 87.84/85.87
& 66.23/66.03
& 56.51/55.54
& 58.46/54.38
& 53.67/52.80
& 74.70/72.36
& 71.44/70.94
& 66.08/63.82
& 11.19/10.06
\\

& RPSRCA
& 53.54/48.64
& 90.73/89.58
& 67.11/66.94
& 63.43/63.11
& 59.85/56.54
& 57.63/56.81
& 79.11/77.41
& 70.63/70.27
& 67.75/66.16
& 8.75/7.75
\\

\rowcolor{NAWIFBlue}
&
\textbf{NAWIF}
& 70.52/\textbf{59.91}
& \textbf{92.39/91.50}
& \textbf{72.65/72.60}
& \textbf{69.89/69.88}
& 59.97/\textbf{58.10}
& \textbf{64.21}/62.60
& \textbf{82.87/81.66}
& \textbf{76.93/76.59}
& \textbf{73.68/71.60}
& \textbf{2.56/1.12}
\\

\bottomrule

\end{tabular}
}

\vspace{1.2mm}

\raggedright
\fontsize{7.1}{8.0}\selectfont
Avg. Rank is calculated across all compared methods;
a lower value indicates better overall performance.

\end{table*}

\subsection{Experimental Setup}
\label{sec:experimental_settings}

The experimental settings are summarized as follows.

\textbf{\textit{Datasets:}}
Eight real-world UCI benchmark datasets~\cite{untoro2020evaluation} are considered: Iris, Wine,
Heart (Statlog), Breast Cancer Wisconsin (Original, BCO),
Breast Cancer Wisconsin (Diagnostic, BCD), Pima Indians Diabetes,
Haberman, and Mammographic Mass. Missing values are imputed using
the median estimated from the corresponding training fold, followed
by min--max normalization fitted only on the training data.

\textbf{\textit{Machine-learning (ML) classifiers:}}
We compare NAWIF with representative ML classifiers: BayesNet, support vector machine (SVM), decision tree (DT), \(k\)-nearest neighbors (\(k\)-NN), AdaBoost, and multilayer perceptron (MLP).

\textbf{\textit{Evidential classifiers:}}
Six evidential reasoning-based methods are further included: DRCM~\cite{xu2013new}, ER Rule~\cite{yang2013evidential}, MRCM~\cite{murphy2000combining}, EDM~\cite{yong2004combining}, BJSM~\cite{xiao2019multi}, and the recent RPSRCA~\cite{deng2024random}, which either construct BPAs through heuristic allocation schemes or restrict belief assignment to a limited set of focal elements.

\textbf{\textit{Implementations:}}
All methods use identical data splits and are evaluated by 100 repetitions
of five-fold cross-validation. Performance is measured by the average
accuracy (ACC) and Macro-F1. To evaluate robustness, Gaussian noise is added to the normalized training and test features, using $\mathcal{N}(0,0.1^2)$ for Haberman and $\mathcal{N}(0,0.4^2)$ for all other datasets.

\subsection{Performance Comparison}

Table~\ref{tab:overall_comparison} reports the comparative results on clean data. NAWIF achieves the highest average ACC and Macro-F1 scores of 87.34\% and 85.47\%, respectively, with the lowest corresponding average ranks of 2.38 and 2.00.

As shown in Table~\ref{tab:overall_comparison_noisy}, NAWIF remains competitive under Gaussian perturbations, attaining the highest average ACC and Macro-F1 of 73.68\% and 71.60\%, respectively, exceeding the corresponding second-best results by 1.04 and 2.31 percentage points. It further achieves the highest Macro-F1 on seven of the eight datasets. These results indicate that negation-aware BPA generation and adaptive evidential discounting effectively mitigate noise-induced overcommitment while suppressing unreliable and uninformative evidence before fusion.

\subsection{Ablation Study and Further Analysis}

As shown in Table~\ref{tab:ablation}, negation-aware BPA generation is the primary contributor to NAWIF. Removing it reduces the average ACC/Macro-F1 from 87.34\%/85.47\% to 83.09\%/81.16\% on clean data, and from 73.68\%/71.60\% to 67.48\%/64.91\% under noise. Evidential discounting provides further gains, with global reliability yielding the largest improvement.

\begin{figure}
    \centering
    \includegraphics[width=0.95\linewidth]{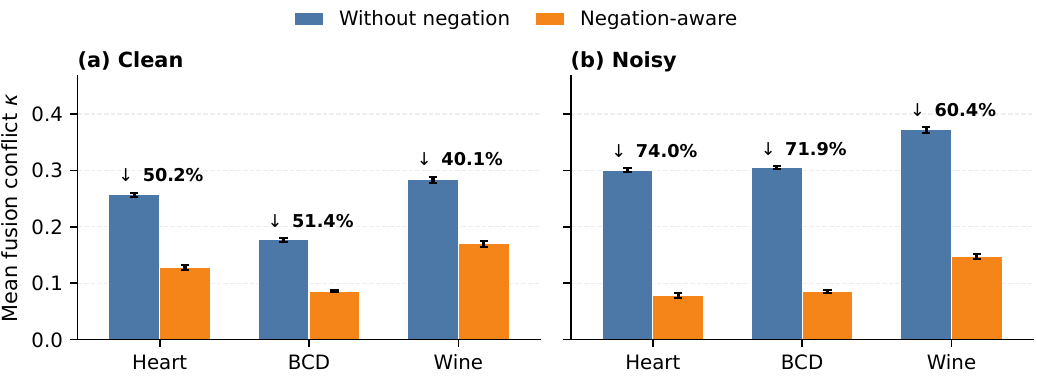}
    \caption{Average combination conflict of different information sources on clean and noisy datasets.}
    \label{fig:placeholder1}
\end{figure}

The benefit of negation-aware modeling is further illustrated in Fig.~\ref{fig:placeholder1}. The non-negation baseline assigns belief only to singleton hypotheses and the full set, which may prematurely commit ambiguous evidence to individual classes. In contrast, negation-aware BPA generation represents ambiguity through complementary and composite propositions, consistently reducing conflict in subsequent fusion. This effect is more pronounced under noise, with conflict reductions of 60.4--74.0\% compared with 40.1--51.4\% on clean data, as perturbations increase source ambiguity and conflicting singleton commitments. These results indicate that preserving ambiguity through composite hypotheses effectively mitigates noise-induced overcommitment.

\begin{table}[t]
\centering
\caption{Ablation study of NAWIF components on the datasets in Table~\ref{tab:overall_comparison}. Neg., R, and I denote negation-aware operation, reliability, and information volume, respectively.}
\label{tab:ablation}

\fontsize{7.5}{8.5}\selectfont
\setlength{\tabcolsep}{9.0pt}
\renewcommand{\arraystretch}{0.86}

\begin{tabular}{lccccc}
\toprule
\textbf{Variant}
& \textbf{Neg.}
& \textbf{R}
& \textbf{I}
& \textbf{ACC}
& \textbf{Macro-F1}
\\
\midrule

w/o Neg
&
& \checkmark
& \checkmark
& 83.09 & 81.16
\\

Neg
& \checkmark
& 
& 
& 87.02 & 85.10
\\

Neg-R
& \checkmark
& \checkmark
&
& 87.28 & 85.31
\\

Neg-I
& \checkmark
&
& \checkmark
& 86.65 & 84.81
\\

\rowcolor{NAWIFBlue}
\textbf{NAWIF}
& \checkmark
& \checkmark
& \checkmark
& \textbf{87.34}
& \textbf{85.47}
\\

\bottomrule
\end{tabular}
\end{table}

\section{Conclusion}
This paper proposes NAWIF, a negation-aware weighted evidential fusion framework that combines source-level negation-aware BPA generation with adaptive discounting based on source reliability and evidential uncertainty. Experiments on eight UCI datasets demonstrate state-of-the-art average performance, with particularly pronounced gains under Gaussian noise. Further analysis shows that negation-aware BPA modeling substantially reduces fusion conflict by exploiting composite belief structures to better represent ambiguity and uncertainty.
\bibliographystyle{IEEEtran}
\bibliography{refs.bib}

\end{document}